\documentclass[lettersize,journal]{IEEEtran}
\pdfoutput=1
\usepackage{amsmath,amsfonts}
\usepackage{algorithmic}
\usepackage{algorithm}
\usepackage{array}
\usepackage[caption=false,font=normalsize,labelfont=sf,textfont=sf]{subfig}
\usepackage{textcomp}
\usepackage{stfloats}
\usepackage{url}
\usepackage{verbatim}
\usepackage{graphicx}
\usepackage{cite}
\usepackage{multicol}
\usepackage{booktabs}
\usepackage{url}
\usepackage{hyperref} 
\usepackage{etoolbox}

\title{\begin{flushright}
\vspace{+2.0cm}
\textbf{MapInWild} \\ 
\vspace{+1.0cm}
\LARGE A Remote Sensing Dataset to Address the Question \\
What Makes Nature Wild
\vspace{+3.0cm}
\end{flushright}}

\author{\small\textbf{\begin{flushleft}\vspace{-9cm}
  BURAK EKIM,\thanks{Burak Ekim and Michael Schmitt are with the Institute of Space Technology and Space Applications, Department of Aerospace Engineering, University of the Bundeswehr Munich, Neubiberg, Germany, \\ (burak.ekim,michael.schmitt)@unibw.de}
  \and
  TIMO T. STOMBERG,\thanks{Timo T. Stomberg and Ribana Roscher are with the Institute of Geodesy and Geoinformation, University of Bonn, Germany, \\ (timo.stomberg, ribana.roscher)@uni-bonn.de}
  \and
  RIBANA ROSCHER,
  \and
  MICHAEL SCHMITT\vspace{+6cm}\end{flushleft}}
}

\begin{document}

\markboth{Accepted for inclusion in a future issue of the IEEE GEOSCIENCE AND REMOTE SENSING MAGAZINE}
{\MakeLowercase{\textit{Ekim et al.}}}

\maketitle

\section{INTRODUCTION}

The advancement in deep learning (DL) techniques has led to a notable increase in the number and size of annotated datasets in a variety of domains, with remote sensing (RS) being no exception \cite{review}. Also, an increase in earth observation (EO) missions and easy access to globally available and free geodata have opened up new research opportunities. Although numerous RS datasets have been published in the past years \cite{schmitt_nodata, bighearth, sen12ms, so2sat, fair1m}, most of them addressed tasks concerning man-made environments such as building footprint extraction and road network classification, leaving the environmental and ecology-related sub-areas of remote sensing underrepresented. Nevertheless, environmental protection has always been an important topic in the RS community, with RS being a useful tool to support conservation policies and strategies combating challenges such as deforestation and loss of biodiversity \cite{biodiversity, sdg_persello, wildlife_tuia}. Thus, in this paper, to meet the pressing need to better understand the nature we are living in, we introduce a novel task of wilderness mapping and advertise the MapInWild dataset \cite{miw_igarss_ekim} -- a multi-modal large-scale benchmark dataset designed for the task of wilderness mapping from space. 

Automated classification of image data has a long tradition in EO. In this community, the classification task can be addressed in different ways, most notably as a scene- or patch-wise classification in contrast to pixel-wise classification. While in scene classification full scenes are assigned by the classifier with single or multiple class labels, in pixel-wise classification (usually called semantic segmentation by the computer vision community) the task outputs densely-annotated prediction maps on a pixel scale by separating the input into distinct and semantically coherent segments.

In general, image classification approaches are either based on hand-crafted feature engineering and subsequent machine learning models, or feature learning incorporated into the machine learning model in the form of deep neural networks. In the RS area, the long-standing tasks of scene classification and semantic segmentation have been approached in a variety of settings \cite{survey_rsic}, ranging from metric learning \cite{metric_learning} to multi-task learning \cite{schmitt_mtl}. While some methods frame the RS-related tasks within the context of perturbation-seeking generative adversarial networks \cite{pertubration}, some others made use of uncertainty estimation applied to deep ensembles \cite{lang_uncertainty} and self-attention context networks under adversarial attacks \cite{selfattentioncontext}.

\begin{figure}
  \centering
  \includegraphics[scale=0.75]{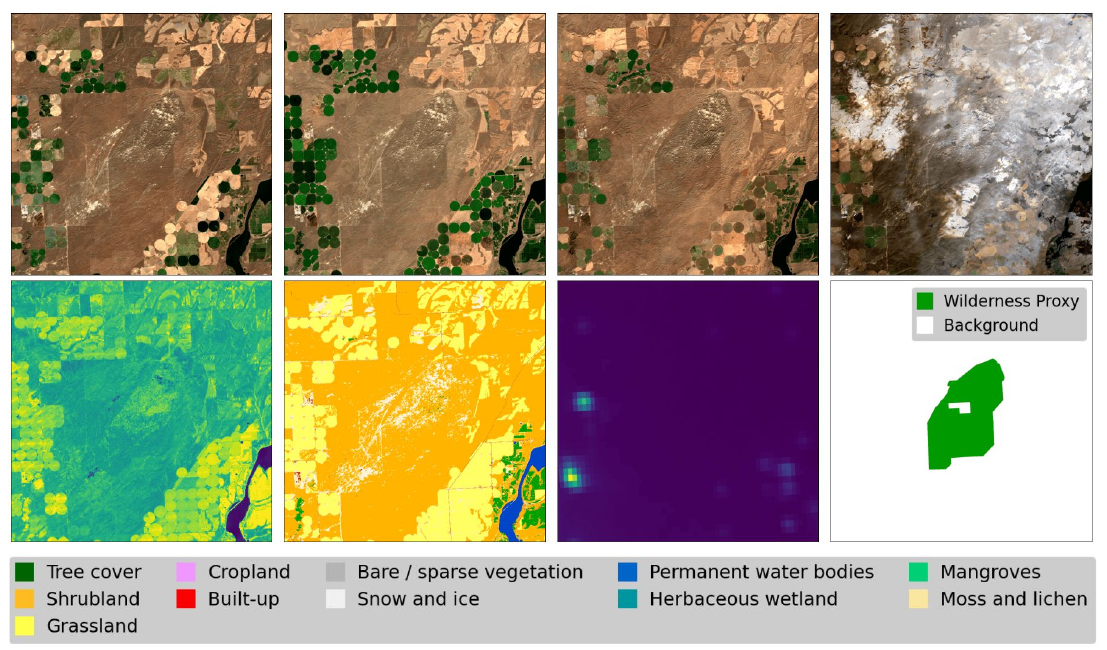} 
    \caption{A wilderness sample from MapInWild dataset (Juniper Dunes, USA. WDPA ID: 555556115). The first row shows the Sentinel-2 images of four seasons, from left to right: spring, summer, autumn, and winter. Second row, same order: Sentinel-1 image, ESA WorldCover map, VIIRS Nighttime Day/Night band, and WDPA annotation. For visualization purposes, the VV and VH bands of the Sentinel-1 image are treated as Red and Green bands and only visible bands (Red: B4, Green: B3, Blue: B2) in the Sentinel-2 image are used. The ESA WorldCover map legend is given below the figure.}
\label{sample_images}
\end{figure}

The success of DL models comes at the expense of decreased interpretability, which means the ability to understand the decision process of the model and the reason why a specific outcome was derived. This is mainly caused by their formation of hundreds of successive layers, leading to a high number of parameters. In recent years, several studies addressed the lack of explainability and interpretability of DL models and proposed methods to overcome this challenge. Model-agnostic approaches are independent of the used model and can be applied post-hoc. Popular approaches are, for example, occlusion sensitivity maps (OSM \cite{zeiler_fergus}) that observe the change in the output while systematically occluding small parts of the input, local interpretable model-agnostic explanations (LIME \cite{lime}) that approximate the model with smaller models with fewer parameters and gradient-weighted class activation maps (GradCAM \cite{gradcam}) that combine the activations maps in a CNN with class-specific gradients. Besides model-agnostic approaches, model-specific approaches are tailored to specific models. They mainly use parts of the model, such as the weights, to analyze the decision process or to interpret the model outcome. One example is ASOS, which analyzes the activation maps and utilizes the concept of occlusion sensitivity in the activation space.

The task of wilderness mapping is interesting in a two-fold way: First of all, wilderness areas are an essential element of the natural environment and provide native habitat for many species, which are often endangered. Thus, the very task itself is important in the context of conservation and environmental protection. Secondly, though, the term wilderness is of philosophical nature and comparably ill-defined from a mathematical/technical standpoint \cite{bastmeijer, stomberg_2022}. 

Explainable machine learning methods can be used to address the task of wilderness mapping. The objective of these methods is to uncover patterns in the decision-making process of the mapping task with the goal of improving the understanding of what makes nature wild. Explainable machine learning approaches, which are used for the discovery of new scientific knowledge such as the exploration of wildlife characteristics, combine interpretation tools that present complex processes as in neural networks in a human-understandable space with domain knowledge to derive explanations \cite{roscher_explain}.

Taking the machine learning perspective further, we claim that training models for wilderness mapping are not only an ideal test-bed for methodical developments in explainable ML but also in other sub-fields such as weakly supervised learning. Thus, the dataset will not only cater to the environmental remote sensing community but also address technical audiences beyond that. 

\section{WILDERNESS MAPPING}
Antrophonegic pressure (i.e., human influence) on the environment is the largest single cause of loss of biological diversity \cite{disturbance_habitat} and the conservation of ecological dynamics and biodiversity is now more important than ever before. With the ever-increasing prominence of the need for environmental conservation, the world is in desperate need of environmental monitoring and quantifying the degree of human influence on the environment. 

While there is a lot of work being done in fields such as unsupervised, semi-supervised, and weakly supervised machine learning, most of the established methods still rely on the supervised learning paradigm, rendering the need for carefully annotated training data. In the context of environmental remote sensing, this can be achieved for rather well-defined target classes such as ``water" or ``forests", but becomes a significant challenge if a less technical interpretation of the environment is the desired goal. As an example, many researchers in the past have presented rule-based approaches to study concepts such as naturalness \cite{allan_etal, ekim_naturalness}, human influence \cite{sanderson_etal}, or wilderness \cite{stomberg_junglenet}. All these approaches are carefully designed by domain experts and cannot be validated against actual, biophysically measurable ground truth. The term ``wilderness`` is philosophical rather than technical in its nature and subject to many different definitions \cite{bastmeijer} with the definition of the European Commission being \cite[p. 10]{eu_wilderness}:

\begin{quote}
``A wilderness is an area governed by natural processes. It is composed of native habitats and species, and large enough for the effective ecological functioning of natural processes. It is unmodified or only slightly modified and without intrusive or extractive human activity, settlements, infrastructure or visual disturbance."
\end{quote}

This definition assigns four central ecological aspects to the concept of wilderness: 1) naturalness, 2) undisturbedness, 3) undevelopedness, and 4) scale. While scale can easily be defined in a mathematical sense, e.g. by defining so-called minimum mapping units, naturalness refers to ecosystems functioning in a natural way and can thus hardly be measured in a technical sense, but is subject to judgment by ecology experts based on in situ observations. Undisturbedness and undevelopedness, however, are concepts that can potentially be observed with the use of remote sensing technologies. As per the European Commission, undevelopedness refers to the absence of ``habitation, settlements or other human artefacts such as power lines, roads, railways, fences [that] may hinder ecological processes directly or by promoting the likelihood of human interference" \cite[p. 12]{eu_wilderness}.

While the task of wilderness mapping could be considered as a special case of land cover mapping, there is a subtle difference to be noted that appear from the ill-posed definition of the term \emph{wilderness}. Since there are nomenclatures with precise class definitions to follow for conventional land cover mapping, it is relatively easier to decide what class to assign to the instance of interest, especially in the presence of domain knowledge. However, given the vague definition of the term, we argue that wilderness is not an instance to map, but a concept to discover. 

It is undeniable that wilderness areas pose utmost significance given their role in harboring species and habitats which are indisputable elements of the environment. This feature of the term wilderness gives a rise to incorporating explainable machine learning methods with the purpose of interpreting the decisions made by the DL.

Wilderness has no biophysical basis, and thus cannot be easily delineated or categorized. Instead, wilderness is a cultural concept \cite{role_wilderness_} and the associated definition of wilderness changes over time. Moreover, changing policy and management strategies make the identification and protection of wilderness compelling. There exists a diverse set of explanations of the word wilderness, most of which are engaged with describing it in a philosophical way, which might seem vague from a technical point of view. Besides, understanding wilderness -a key element of nature- poses great importance since the protection of nature can be fostered  by better defining its ill-defined elements. Consequently, we adopt a bottom-up approach and address the ambiguity in nature's vital elements (e.g., wilderness) to better protect the environment we live in. Ultimately, we raise the rhetorical question "What makes nature wild?" and invite the community to participate in finding possible solutions to  the open challenges mentioned above to get closer to discovering the concept of wilderness.

\section{THE MAPINWILD DATASET}

\subsection{CURATION}
The cloud storage and computing facilities of Google Earth Engine (GEE) \cite{gee} are utilized during the dataset curation phase by keeping the reproducibility aspect in mind (i.e., defining fixed seed values to random number generators and using definitive time frames for image queries). Given the complexity of the task being addressed, MapInWild consists of distinct sets of geo-sensors that aim to leverage complementary information about the natural environment: Sentinel-1, Sentinel-2, ESA WorldCover, Visible Infrared Imaging Radiometer Suite (VIIRS), and World Database of Protected Areas (WDPA) polygons where a sample from MapInWild is shown in Figure \ref{sample_images}. WDPA polygons are comprised of feature classes designating protected areas and formed by the United Nations Environment Programme’s World Conservation Monitoring Centre (UNEP-WCMC) and the International Union for Conservation of Nature (IUCN) - World Commission on Protected Areas (WCPA) \cite{iucn_wdpa}. The complete list of WDPA classes is given in Table \ref{table:WDCAT}. From the antrophonegic pressure point of view in which the variations in the diversity are explained along human influence gradients, we reformulate these discrete sets of areas on a continuous spectrum, where the distinction between areas is relatively ill-defined. On the grounds of the observation that human presence increases going from category Ia to category II, we raise the question at what point of the human presence continuum an area should be considered a wilderness area. Supported by this theoretical research question, we argue that category Ia and category II could be beneficial proxies in interpreting and explaining the concept of wilderness. Ultimately, we form the MapInWild dataset with the following three WDPA classes: Strict Nature Reserve (category Ia), Wilderness Area (category Ib), and National Park (category II). 

\begin{figure}
  \centering
  \includegraphics[scale=0.54]{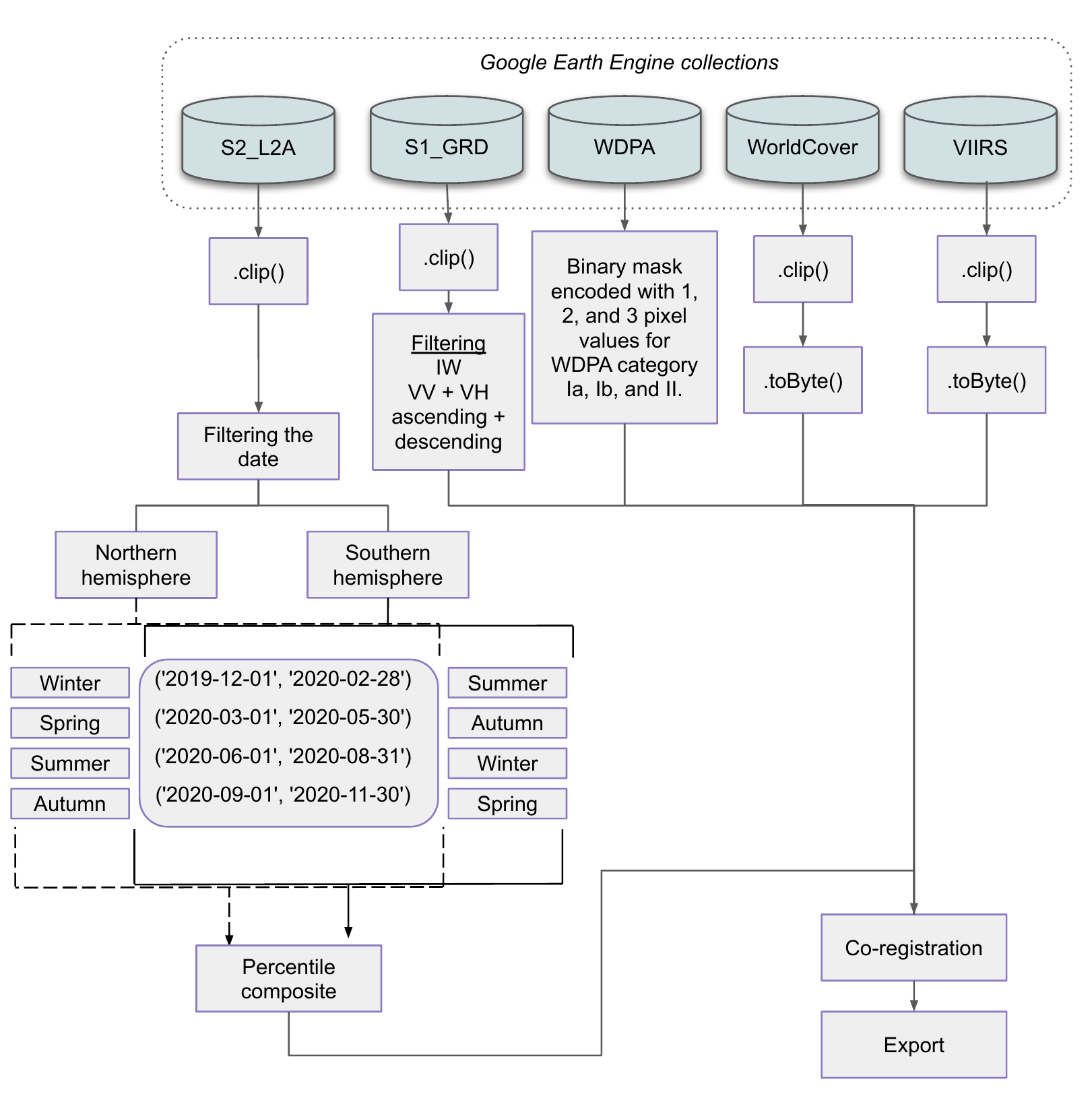}
    \caption{Curation workflow: Querying and filtering the geodata sources in MapInWild dataset. L2A: Level-2A, GRD: Ground Range Distance, VIIRS: Visible Infrared Imaging Radiometer Suite.}
\label{export}
\end{figure}

As for the Sentinel-1 image, a ground-range-detected product with a resolution of 10 meters is used. Sentinel-1 images contain VH and VV polarizations and use interferometric wide as a swath mode. The Level-2A Sentinel-2 image contains 10 spectral bands (B2, B3, B4, B5, B6, B7, B8, B8A, B11, B12) and is subject to mosaicking so as to mitigate the effect caused by clouds. We stack imagery of any area of interest (AOI) for a defined period of time in the year 2020, and for each band in each pixel, we select the value defining the 25 \% percentile regarding reflectivities observed in that time period. We empirically found that this way of compositing results in a more informative image than median compositing. In order to further widen the scope of the dataset, we provide four seasons of Sentinel-2 images for every AOI. By adding a multi-temporal value to the dataset by including all seasons for every AOI, we aim to further facilitate the investigation of wilderness mapping in a time-series manner. Further, rather than defining a single season period when querying the seasons, we take hemispheres of the AOI into account and define the season periods accordingly (i.e. northern hemisphere and southern hemisphere). We use the remaining sources (ESA WorldCover and VIIRS) in their original forms and leave end-users with more flexibility. We performed re-projection and scaling to have the corresponding pixels across geodata sources in the same scale and projection which uses the rubber-sheet algorithm as a registration method. The curation strategy is illustrated in Figure \ref{export}.

\subsection{SAMPLING STRATEGY AND EXPORTING}

\begin{figure}
  \centering
  \includegraphics[scale=0.5245]{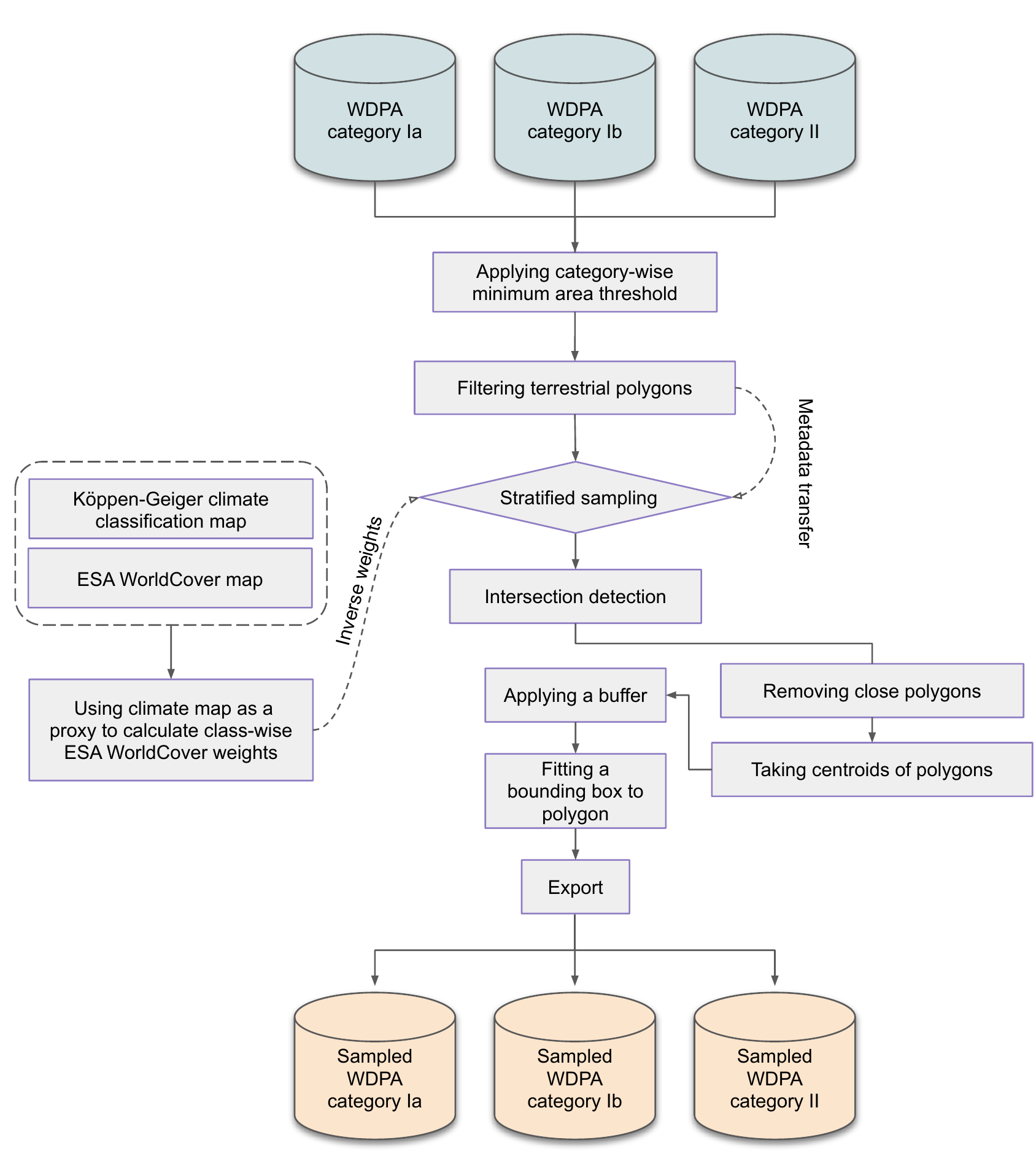}
    \caption{Sampling workflow: The guided stratified sampling of the WDPA polygons. The WDPA polygons that are varying in size and shape are filtered through a semi-automated sampled process that is guided by the climate zones and the land cover types.}
\label{sample}
\end{figure}

Considering the difficulty of adequately sampling the earth's surface in terms of representability, we develop a guided sampling approach taking the hard constraints of the WDPA polygons into account. Prior to filtering images, sampling of WDPA polygons take place. Merely using the original WDPA polygons is not ideal because of the uneven distribution across continents and disparity in polygons (both in terms of volume and size). To this end, to ensure the versatility of the dataset and the spatial coverage of its samples, we design a climate map and land cover type-aware semi-automated sampling approach where a sub-sampling of the polygons is performed while improving the spatial coverage of WDPA polygons. The developed approach is illustrated in Figure \ref{sample}. We start the process by applying a class-wise minimum area threshold (using the asset feature \textit{GIS\_AREA}) of 5 km$^2$ to category Ia and category Ib, and 100 km$^2$ to category II. Then, after filtering out the non-territorial areas, the sampling operation takes place where we use a GEE function (namely, \textit{StratifiedSample()}) that allows users to supervise the sampling process with weights which can be obtained from auxiliary data. For our case, we use the Köppen-Geiger climate classification map and the ESA WorldCover map as a proxy to maintain guidance over the sampling process. The sampling weights used in the \textit{StratifiedSample()} function are formed by calculating the class-wise correlation between the WorldCover and climate classification maps (i.e. frequency of each land cover class in each climate classification map). Consequently, the calculated weights are inversely normalized and row-wise summations are used as sampling weights. Hence, oversampling of underrepresented polygons (and vice-versa) is performed by following the observation that the representability of protected areas could be promoted by taking land cover and climate zones into account. Further, the polygons within a certain proximity (30km$^2$ for category Ia and category Ib, and 50km$^2$ for category II) are removed. Ultimately, the resulting samples are exported for further analysis after fitting a bounding box with the size of 20 km $\times$ 20 km to the center of each polygon. 

After exporting, the images are renamed after the WDPA polygon they contain and cropped into $1920 \times 1920$ pixel patches to remove the re-projection effect occurring near borders. We further add 108 AOIs from populous areas that exhibit human disturbance on Earth in various forms to strengthen the versatility of the dataset. The motivation behind this manual intervention, which increased the number of AOIs in the dataset from 910 to the final number of 1018, is due to the potentiality that the model might overfit the wilderness areas in the form of forested areas (see the ESA WC class distribution in Figure \ref{wc_stat}). A naming convention of \textit{9000000XXX} (where X is the identification number of the patch) is adopted when including the manually-selected AOIs in the dataset. Overall, our dataset has 8144 images (1018 × 8, where 1018 is the number of AOIs in the dataset and 8 is the number of the data sources each AOI contain) with the shape of $1920 \times 1920$ pixels. A sample image from the dataset is given in Figure \ref{sample_images}. 

\begin{table*}
    \centering
    \caption{IUCN protected area categories}
    {\footnotesize
    \begin{tabular}{m{8cm}m{8cm}}
     \toprule
     WDPA Class & Description \\
     \cmidrule(r){1-1}\cmidrule(l){2-2}
     Ia: Strict nature reserve & Strictly protected areas where human presence is strictly limited and controlled. \\

     Ib: Wilderness area &  Slightly modified areas with little human presence in the form of indigenous and local communities. \\

     II: National park  & Functioning ecosystems which subject to tourism through zoning.  \\

     III: Natural Monument or feature  &  Areas where preserving a particular feature hosting a cultural value is the dominant goal.  \\
 
     IV: Habitat/species management area & Areas where flora species, fauna species, or habitats are aimed to be preserved and/or restored through informed interventions. \\\addlinespace

     V: Protected landscape/seascape & Areas where a distinct value is created by human presence over time. \\

     VI: Protected area with sustainable use of natural resources & Areas where the conservation of natural ecosystems and ecological processes take place. \\
  \bottomrule
    \end{tabular}
    }
    \label{table:WDCAT}
\end{table*}

\subsection{QUALITY ASSESSMENT AND STATISTICS}
With the intention of further improving the usability of the dataset, we conduct a quality assessment of the Sentinel-2 seasons. First, an evaluation platform is prepared and shared with many remote sensing experts. The participants are then asked to annotate the informativeness of each Sentinel-2 season. Next, by performing a statistical analysis that takes the percentage of the votes, we (i) discard the patches with a discard score of higher than 50 \%, (ii) assign a quality score between 0 and 10 to each season, (iii) calculate the single temporal subset of each season using the quality scores calculated in the previous step. In cases where several seasons are assigned with the same quality score, we favor the season summer over the other seasons. In addition to the single temporal subsets, we also make the quality scores publicly available which can be used to form case-specific single temporal subsets.

With this quality assessment, we not only refine the dataset and remove non-informative patches, but also boost the versatility of the dataset by generating quality scores and single temporal subset seasons assigned to each season and patch, respectively. We advise the researchers to make use of the single temporal subset seasons defined for each patch should they be solely interested in the single-temporal information. 

\begin{figure*}
  \centering
  \includegraphics[width=0.9\textwidth]{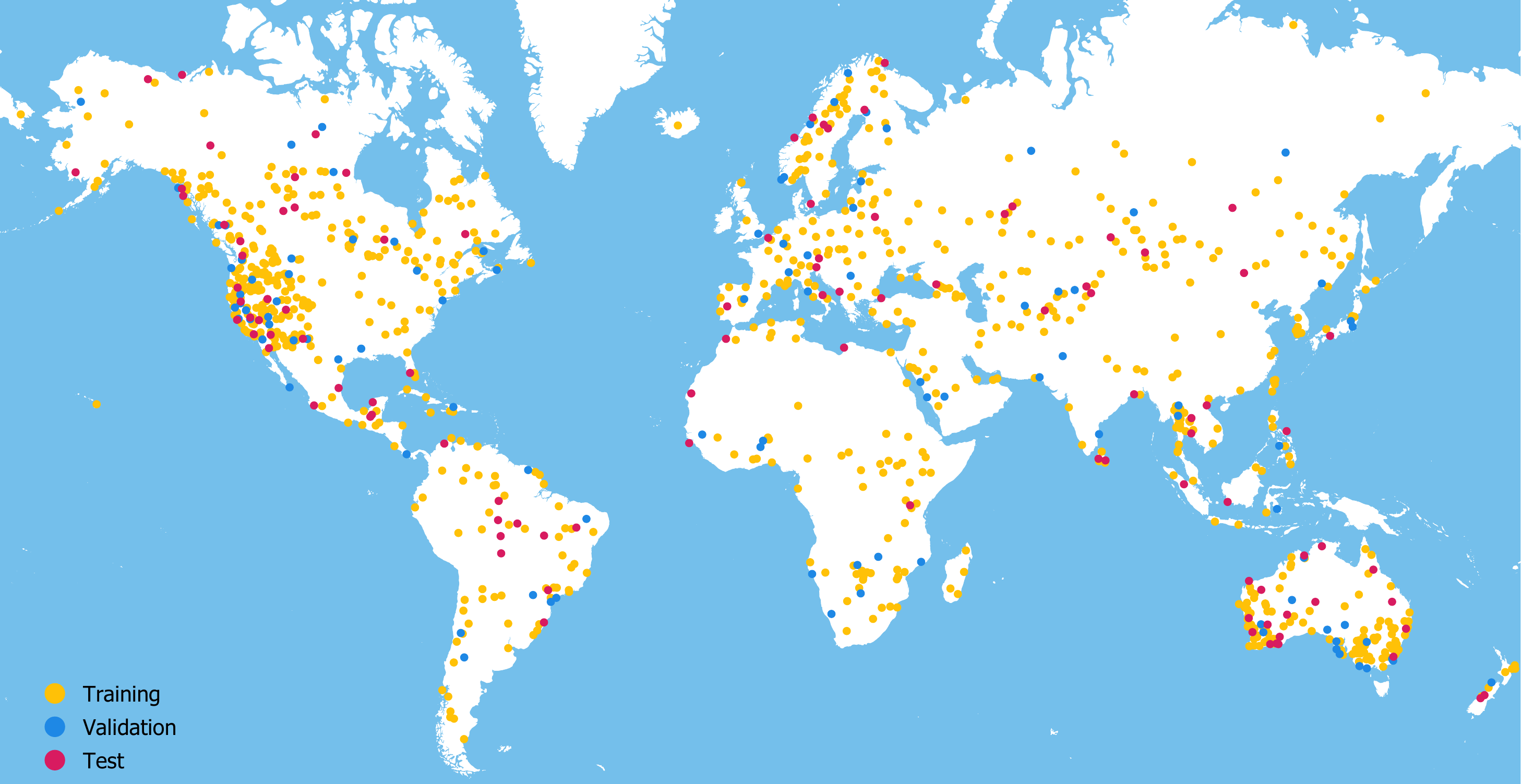}
    \caption{Distribution of the AOIs and training, validation, and test sets across continents. The points represent the center of each study area and are enlarged for better visualization.}
\label{distri}
\end{figure*}

\begin{figure}
  \centering
  \includegraphics[scale=0.22]{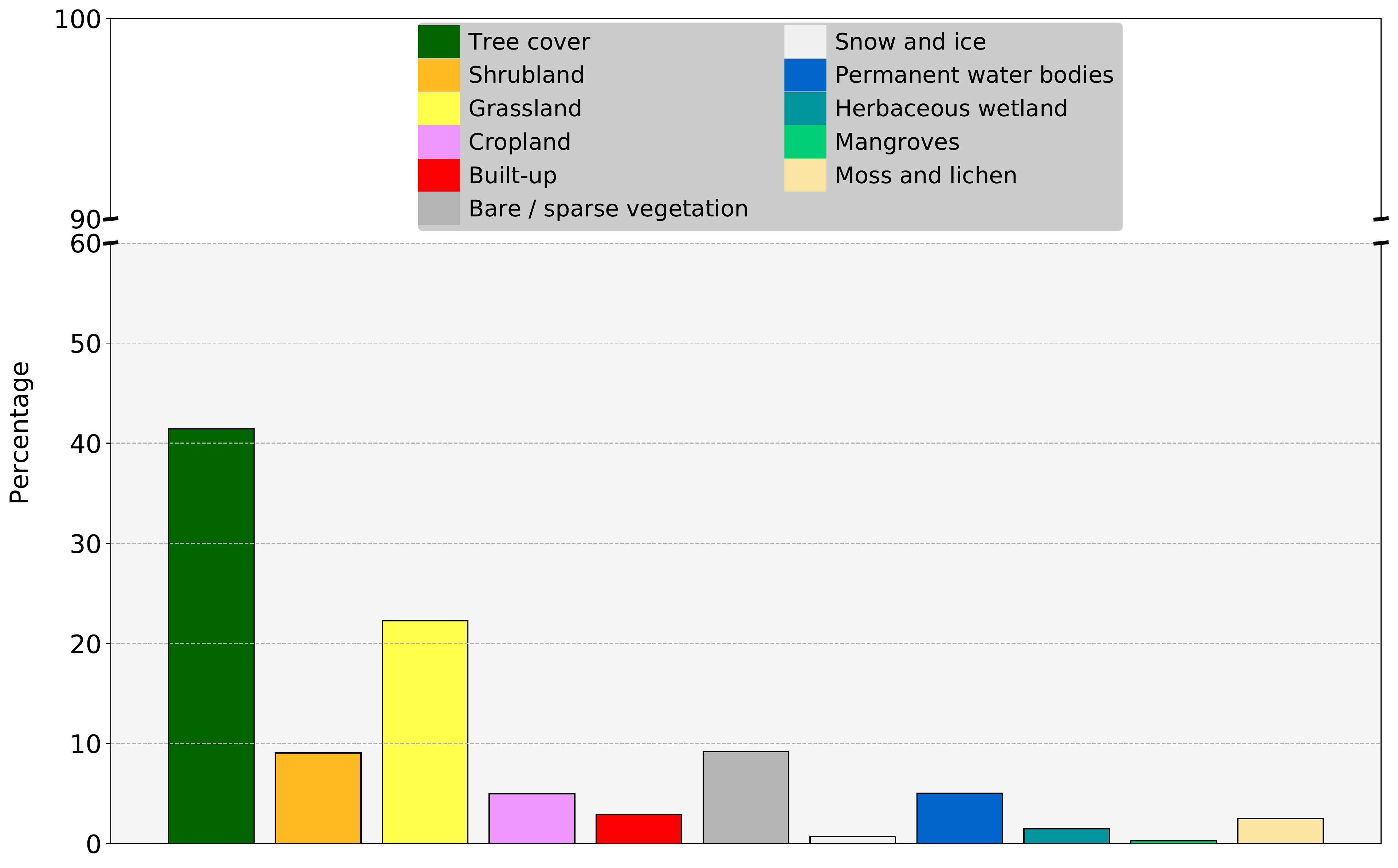}
    \caption{The pixel-level breakdown of final AOIs in terms of land cover. The land cover information is obtained from the ESA WorldCover map existing in the dataset.}
\label{wc_stat}
\end{figure}

The final distribution of the AOIs sampled from the WDPA polygons is given in Figure \ref{distri}. Figure \ref{wc_stat} shows the percentage of ESA WorldCover classes falling in the AOIs.

\section{BASELINE EXPERIMENTS}
In the experiments below, we aim at discovering the concept of wilderness in two different ways; semantic segmentation and scene classification with a focus on sensitivity analysis. Both experiments work towards providing initial efforts in understanding wilderness. For both experiments, out of 1018 patches, we used 80 \%, 10 \%, and 10 \% of them for training, validation, and test sets, respectively. The patch IDs of the sets are made publicly available and the distribution of the sets is shown in Figure \ref{distri}. For the semantic segmentation experiment, we unify the three protected area classes in the annotation sources to form binary wilderness masks. 

\subsection{SEMANTIC SEGMENTATION}
\subsubsection{EXPERIMENTAL SETUP}

\begin{figure}
  \centering
  \includegraphics[scale=0.5]{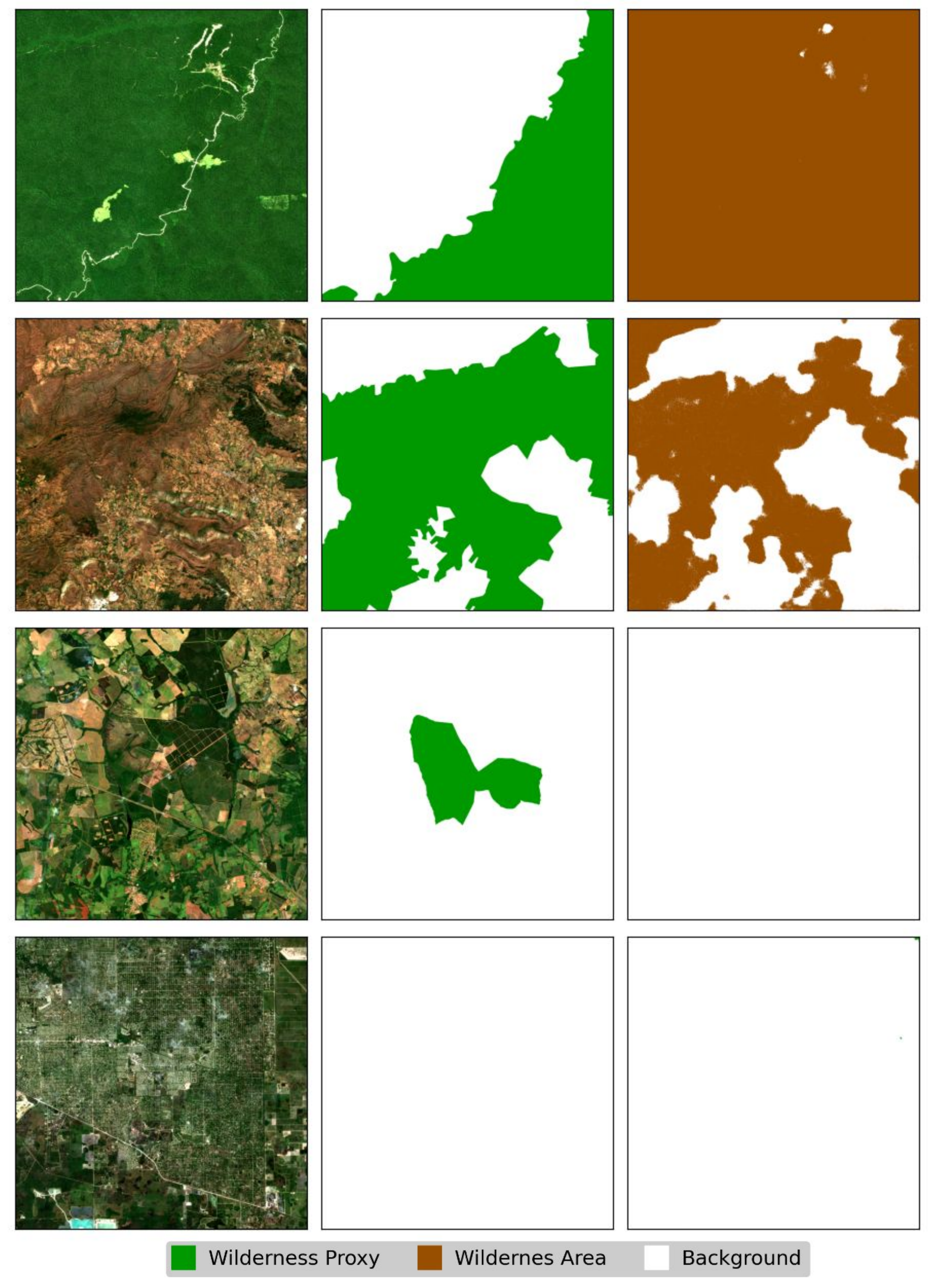}
    \caption{Baseline results for the test images with the IDs 351819, 39516, 19442, 900000068. Row-wise, from left to right: Input image, wilderness proxy, and model's output. Input images are the single temporal subsets of the Sentinel-2 images for the corresponding IDs.}
\label{semseg_outputs}
\end{figure}

For this experiment, we adopt the U-Net architecture \cite{ronneberger_2015} with the ResNeSt 14d \cite{resnest}. The model consists of approximately 8 million parameters. As an optimization algorithm, we use Adam with beta values of (0.9, 0.999) and with an epsilon value of $10^{-8}$. We select the learning rate as 0.0001 and adopt the cosine annealing learning rate schedule with the maximum number of iterations set to 10. We use the dice loss to quantify the cost value and set the batch size as 32. During the training process (e.g., on-the-fly), we randomly extract patches with the shape of $512 \times 512$ pixels from the input images and feed them to the model. We use the three-band (B4, B3, B2) Sentinel-2 image queried with the single temporal subset seasons and utilize ImageNet pre-trained weights during the feature extraction phase. Furthermore, we configure mixed-precision training to shorten the training and inference time and decrease the required amount of memory. We train the model on an NVIDIA Tesla V100 SCM3 32GB. The training process is terminated with an early stopping callback with patience (number of epochs without improvement in the metric observed) of 8. The achieved overall pixel-wise classification accuracy, intersection over union (IoU) score, and F1 score are shown in Table \ref{tab:qaulitative}. The quantitative results are shown in Figure~\ref{semseg_outputs}.

\begin{table}
\centering
\caption{Qualitative Results (\%): Baseline Model for the Semantic Segmentation Task}
\begin{tabular}{l|r|r|r|r}
\toprule
Model & Backbone & IoU & Accuracy & F1  \\\hline
U-Net & Resnest 14d & 69.1 & 76.31 & 81.73 \\\hline
\end{tabular}
\label{tab:qaulitative}
\end{table}

\subsection{SCENE CLASSIFICATION AND SENSITIVITY ANALYSIS}

In this experiment, we perform scene classification with a subsequent sensitivity analysis. To this end, we use the single-temporal subset of the Sentinel-2 data and extract tiles with a height and width of 256 pixels each. The tiles are extracted in such a way that they are located exclusively either within or without the WDPA regions without overlapping. Thus we get 24,101 tiles within WDPA regions (label 1) and 17,795 tiles outside of WDPA regions (label 0). The proportions of the training, validation, and test set remain approximately the same.

We perform Activation Space Occlusion Sensitivity (ASOS) by \cite{stomberg_2022} to create high-resolution sensitivity maps of our test samples to assess whether specific regions are sensitive toward wilderness or non-wilderness. The authors provide neural network architecture, a specific training procedure, and the approach for sensitivity analysis. We proceed in the same way and refer to their work for a detailed description of the methodology.

\begin{figure*}
  \centering
  \includegraphics[width=0.99\textwidth]{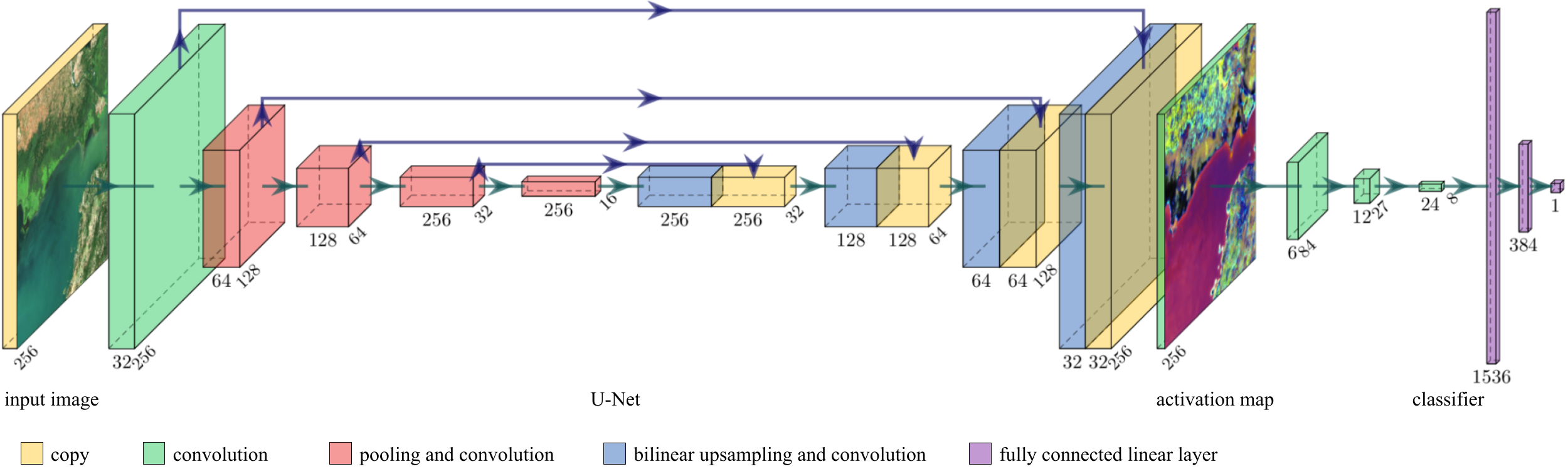}
  \caption{For the sensitivity analysis, we use a neural network consisting of a modified U-Net and an image classifier. The activation map at the interface is essential for determining the sensitivities. \tiny{The figure is taken from \cite{stomberg_2022} and slightly adapted.}}
\label{fig:nn_architecture}
\end{figure*}

The neural network consists of a modified form of the U-Net by \cite{ronneberger_2015} and a simple classifier network, shown in Figure~\ref{fig:nn_architecture}. The activation map at the interface of these two networks has three channels and is the same size as the input image. The model has about 2 million parameters.

\subsubsection{CLASSIFICATION}

We run the same training procedure as \cite{stomberg_2022} and use a maximum learning rate of 0.01, a weight decay of 0.0001, and a batch size of 32. We train the model for 50 epochs on an NVIDIA Quadro RTX 4000 (8 GB GDDR6) for about six hours.

The achieved overall classification accuracies are 92~\% for the training set, 82~\% for the validation set, and 74 \% for the test set. The confusion matrix of the test set is shown in Table~\ref{tab:confusion_matrix}.

\begin{table}
    \centering
    \caption{\label{tab:confusion_matrix}Confusion matrix of the test dataset for the classification task. Percentages are given in brackets}
    \begin{tabular}{l|rr}
                 & prediction: &       \\
                 & non-wild    &  wild \\ \hline
        label:   &             &       \\
        non-wild &       1,062 [62.65\%]&   633 [37.34\%] \\
        wild     &         439 [17.79\%] & 2,028 [82.2\%]        
    \end{tabular}
\end{table}

\subsubsection{SENSITIVITY ANALYSIS}

\begin{figure}
  \centering
  \includegraphics[width=0.90\columnwidth]{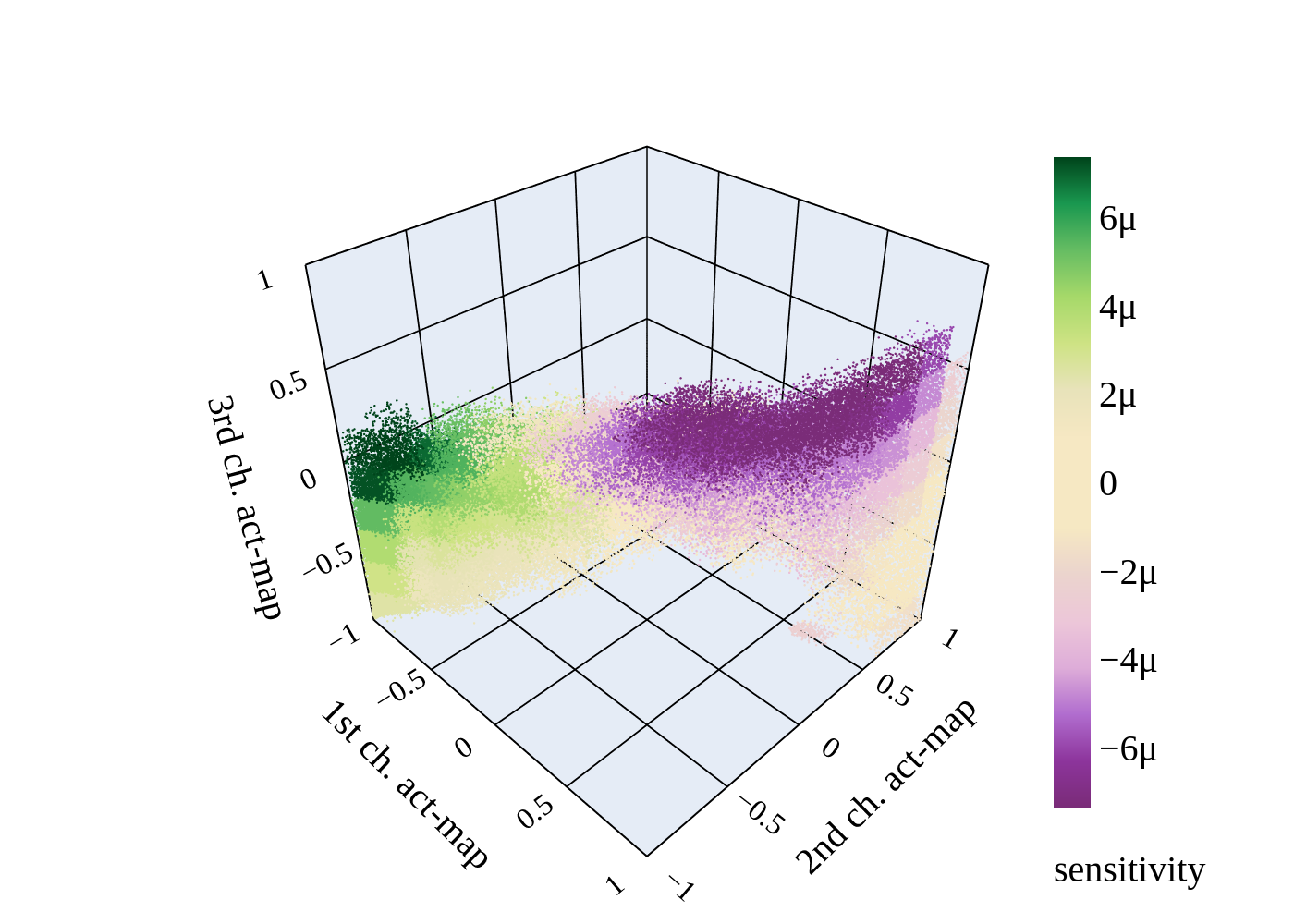}
  \caption{Activation space of the sensitivity analysis. Each axis corresponds to the values of the respective channel in the activation maps. The greener the vectors, the more they are sensitive to wilderness characteristics; the more violet, the more sensitive to non-wilderness characteristics. Low-density areas are not included in the sensitivity mapping and the corresponding vectors are not shown here.}
\label{fig:activation_space}
\end{figure}

\begin{figure}
  \centering
  \includegraphics[width=0.99\columnwidth]{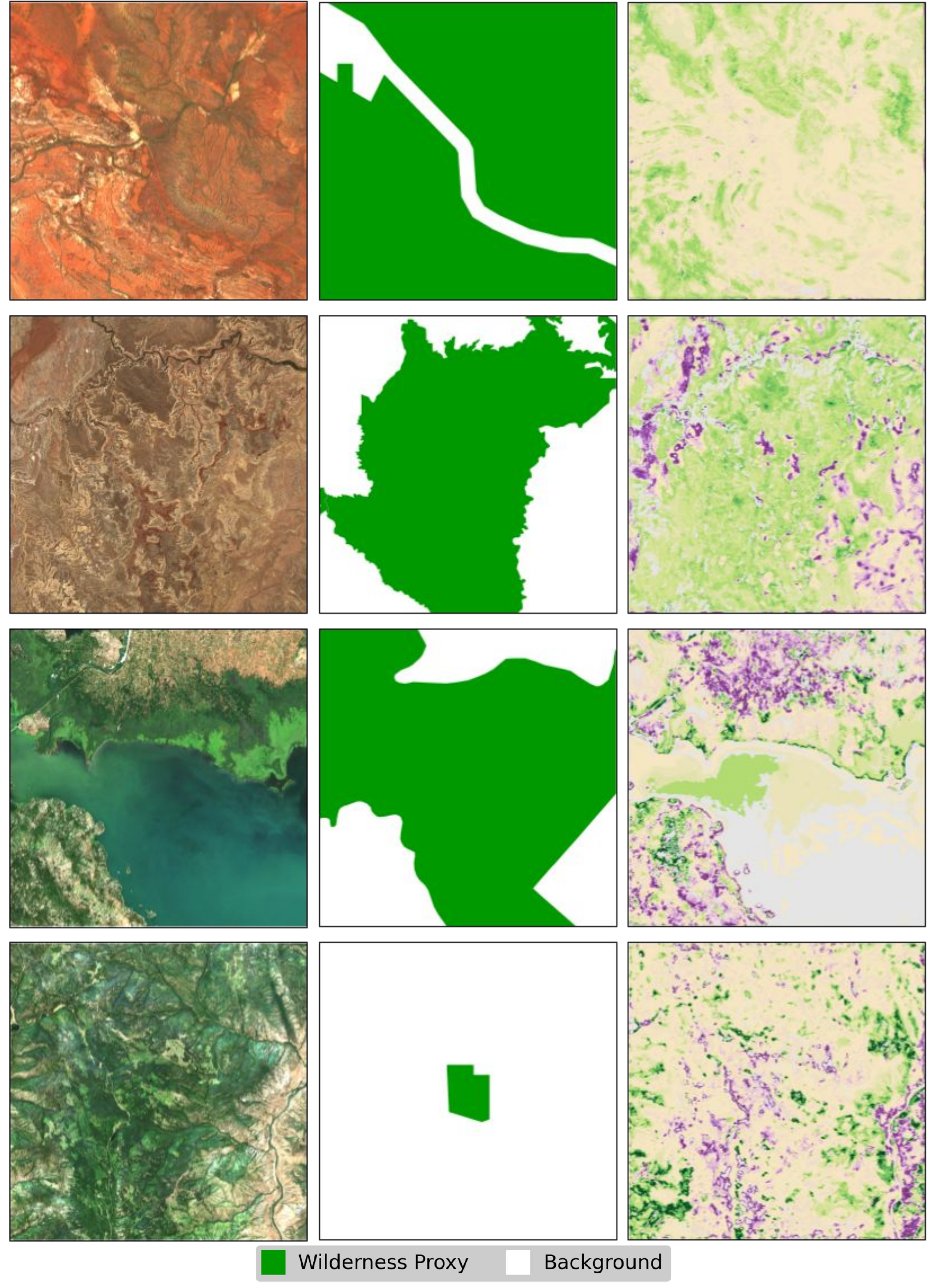}
  \caption{Baseline results for the test images with the IDs 64113, 374681, 16385, 18415. Row-wise, from left to right: Input image, wilderness proxy, and sensitivity map. The color scale of the sensitivity maps is given in Figure~\ref{fig:activation_space}. Areas with a low density in the activation space are not mapped to sensitivities and are colored grey in the sensitivity maps.}
\label{fig:samples_asos}
\end{figure}

After the model has been trained, we run the ASOS analysis on all correctly classified training samples. For this, specific pixels are occluded in the activation maps by setting them to zero. This changes the classification scores and the deviations are a measure of the sensitivities of the occluded parts.
The occlusions are determined from the activation space in which each of the three axes corresponds to one of the three channels of the activation maps. Pixels that are close to each other in the activation space are occluded simultaneously. The resulting sensitivities are visualized in Figure~\ref{fig:activation_space}. Low-density areas are not included in this mapping. Knowing the sensitivities, we can predict sensitivity maps for any input image. A selection of the test samples is shown in Figure~\ref{fig:samples_asos}.

\section{DISCUSSION}

With the semantic segmentation experiment, we study the concept of wilderness on a pixel scale. The model used in this experiment outputs densely-predicted segmentation maps for a given test image with the shape of $1920 \times 1920$, as shown in Figure \ref{semseg_outputs}. The first sample shows parts of the Rio Novo National Park where the park is seen as divided in  two by a river. We see that the model predicts the whole patch (except for the deforestation areas) as wilderness while the WDPA proxy contains annotation only on the right side of the river. The second sample shows the Phu Pha Man District in Thailand. Here, the proxy loosely aligns with the models' output. The output segmentation map here filters out the areas with a human influence (i.e., a sub-urban area surrounded by cultivated lands). The third sample is located at Santa Bárbara Ecological Station in Brazil. Here, the segmentation map does not include any annotation. This could be explained by the observation that this area exhibits human influence in the form of roads and arable lands which clearly does not comply with the "undisturbedness" and "undevelopedness" states of the wilderness areas. The last sample shows Cape Coral, Florida. This is one of the samples we have manually added to the dataset (notice the 9 at the beginning of the ID). This sample shows that the model does not annotate an urban area as wilderness although there are golf courses, inner-city parks, football courts, and forest areas on the perimeter. This sample also set forth the ability of the model to capture the "disturbedness" and "developedness" of an area. All things considered, this experiment exhibits that the model is competent to discover wilderness areas on a pixel scale. While doing so, the model provides a more viable annotation of wilderness areas as an output. With this observation, we go a step further and analyze the models' output with the use of explainable ML techniques to take a deeper look at the models' decisions.    

Applying ASOS to MapInWild dataset we demonstrate an example task for explainable ML. Training a neural network on classification, we can predict high-resolution sensitivity maps shown in Figure~\ref{fig:samples_asos}. The first sample shows parts of the Australian national park Karijini. It is centrally split by a mine and a railway. The human influence on the surroundings appears to be so small that the model does not highlight them as such. The second sample is located at Sid's Mountain Wilderness Study Area in Utah, United States. The model detects some parts which are mainly outside the study area as non-wild. The inner parts of the study area seem to have wild characteristics. The third sample shows the Skadar Lake National Park in Montenegro and Albania. At the top, agricultural fields reach inside the national park, which is detected by the model. The lake itself has mainly been not predicted due to high uncertainty. This goes with the MapInWild dataset which has been created by specifically looking for non-water polygons. The fourth sample shows the small Skwaha Lake Ecological Reserve in Canada. This reserve has, according to the model, mixed characteristics. The two regions detected as non-wild are valleys. The left one includes several streets and the right region includes the Trans-Canada Highway 1 along a river.

We provide the experiments given above as an initial effort in addressing the novel task of wilderness mapping on the large-scale and benchmark dataset MapInWild. The task has first been investigated on a pixel scale within the frame of semantic segmentation where the imperfect wilderness annotations have been used as a proxy when mapping the wilderness areas. 
On the one end, there are cases where the learner extrapolates the wilderness area to the entire scene as the wilderness area in the annotation is of alike appearance as the remaining area which is annotated as non-wilderness (the first sample in Figure~\ref{semseg_outputs}). This behavior of the learner exhibits its ability to learn the concept of wilderness from imperfect annotations. On the opposite end, it is seen that when mapping the wilderness areas the learner dampens the presence of wilderness characteristics shown in the proxy. Although there is an annotated wilderness characteristic in the center of the area (the third sample in Figure~\ref{semseg_outputs}) the existing human influence in the surroundings might be negatively affecting the essence and form of the wilderness concept realized and discovered by the learner. The concept realized might contain a learned pattern on the (i) size of a wilderness area, (ii) distance to the nearest area under human influence, and (iii) characteristics of the site in terms of land use and land cover. Similar to this behavior, the learner shows strength in leaving out the resemblance of inner-city parks and forests with a wilderness area in the form of a forested area (the last sample of Figure~\ref{semseg_outputs}). In the middle of the learner's behavior scale, there lies a typical representation of a semantic segmentation task in which the learner approximates the given annotation (the second sample in Figure~\ref{semseg_outputs}). While doing so, the learner struggles with the borders of the given wilderness area which can be explained by the (i) edge-effect \cite{edge_effects} that complicates the segregation of ecological units from the above perspective and (ii) the imperfect annotations used when training the learner.

Later, the learner has been further investigated by employing an interpretable-by-design architecture to study the patterns in the decision-making process. The high-resolution sensitivity maps in Figure~\ref{fig:samples_asos} make it evident that the learner holds a deeper understanding of wilderness that disentangles wilderness from human influence. The maps produces in this experiment provide pixel-level sensitivity information that could be utilized in the process of inferring new scientific insights. 

Although the behavior of the learner can be often explained away with some confidence, in the light of the experimental results explained above, it is still unclear when and why the learner behaves in certain ways. Motivated by this observation, we urge environmental science, conservation, computer science, and remote sensing researchers to study the ambiguity in the ill-defined elements of nature to better monitor, understand, and protect nature, our home.

\section{SUMMARY \& CONCLUSION}
With this paper, (i) we introduce a novel task of wilderness mapping, (ii) and publish MapInWild, a large-scale benchmark dataset curated for the task of wilderness mapping. MapInWild is a multi-modal dataset and comprises various geodata acquired and formed from a diverse set of RS sensors. The dataset consists of 8144 images with the shape of 1920 × 1920 pixels and is approximately 350 GB in size. The images are weakly annotated with three classes derived from the World Database of Protected Areas – Strict Nature Reserves, Wilderness Areas, and National Parks. With MapInWild, on the purpose of deepening our understanding of what makes nature wild, we embark on the complications induced by the ambiguity of the term wilderness and study the vagueness in nature and propose our dataset as a test-bed for machine learning research concerning environmental remote sensing. We are convinced that getting closer to understanding the concept of wilderness is of great value to the community to further bridge the gap between deep learning applied to environmental remote sensing and conservation. Both MapInWild dataset and the code are publicly available at \url{https://dataverse.harvard.edu/dataverse/mapinwild} and \url{https://github.com/burakekim/MapInWild}. 

\section*{ACKNOWLEDGMENTS}
This work was supported by the German Research Foundation (DFG project MapInWild, grant RO 4839/5-1 / SCHM 3322/4-1). The authors acknowledge the computing time granted by the Institute for Distributed Intelligent Systems and provided on the GPU cluster Monacum One at the University of the Bundeswehr Munich.

\end{document}